\documentclass[a4paper,conference]{IEEEtran}
\IEEEoverridecommandlockouts
\usepackage{cite}

\ifCLASSINFOpdf
   \usepackage[pdftex]{graphicx}
   \graphicspath{{../pdf/}{../jpeg/}}
  \DeclareGraphicsExtensions{.pdf,.jpeg,.png}
\else
\fi
\usepackage{amsmath}
\usepackage{mathtools}
\usepackage{url}

\begin{document}
%
\title{Accelerating the creation of instance segmentation training sets through bounding box annotation}

\author{\IEEEauthorblockN{Niels Sayez*}
\IEEEauthorblockA{ICTEAM\\
UCLouvain, Belgium \\
Email: niels.sayez@uclouvain.be}
\and
\IEEEauthorblockN{Christophe De Vleeschouwer* \thanks{*  Part of this work has been funded by the Walloon Region DeepSport project, and by the Brain-be 2.0 DeepSun Belspo project. C. De Vleeschouwer is funded by the Belgian NSF. }}
\IEEEauthorblockA{ICTEAM \\
UCLouvain, Belgium\\
Email: christophe.devleeschouwer@uclouvain.be}
}



%


\maketitle


\begin{abstract}
Collecting image annotations remains a significant burden when deploying CNN in a specific applicative context. This is especially the case when the annotation consists in binary masks covering object instances. Our work proposes to delineate instances in three steps, based on a semi-automatic approach: (1) the extreme points of an object (left-most, right-most, top, bottom pixels) are manually defined, thereby providing the object bounding-box, (2) a universal automatic segmentation tool like Deep Extreme Cut is used to turn the bounded object into a segmentation mask that matches the extreme points; and (3) the predicted mask is manually corrected. Various strategies are then investigated to balance the human manual annotation resources between bounding-box definition and mask correction, including when the correction of instance masks is prioritized based on their overlap with other instance bounding-boxes, or the outcome of an instance segmentation model trained on a partially annotated dataset. Our experimental study considers a teamsport player segmentation task, and measures how the accuracy of the Panoptic-Deeplab instance segmentation model depends on the human annotation resources allocation strategy. It reveals that the sole definition of extreme points results in a model accuracy that would require up to 10 times more resources if the masks were defined through fully manual delineation of instances.  
When targeting higher accuracies, prioritizing the mask correction among the training set instances is also shown to save up to 80\% of correction annotation resources compared to a systematic frame by frame correction of instances, for a same trained instance segmentation model accuracy.
\end{abstract}


%
\IEEEpeerreviewmaketitle

\section{Introduction}
\label{sec:intro}

To extend the use of deep learning to out of mainstream applications, there is a growing need for efficient solutions to annotate images, and in particular to provide dense segmentation masks for objects-of-interest in a scene. As an example in the video industry, the segmentation of players in teamsport scenes helps in game interpretation \cite{CMR_ref0}\cite{CMR_ref1}, including to support autonomous production \cite{CMR_ref2} and intelligent transmission of associated video content \cite{CMR_ref3}\cite{CMR_ref4}\cite{CMR_ref5}. In this paper, we propose to exploit a universal prior-based segmentation model like DEXTR \cite{Man+18} to reduce the human load involved in the creation of the masks needed to train an instance segmentation model. Instance segmentation considers the pixel-wise delineation of all the individual instances of a class of objects in an image.

Our work proposes to split the manual load associated with the annotation of an instance mask in two parts. The first one consists in defining the extreme points of the instance. Those points are then provided as input to a universal model trained to automatically segment objects in its input bounding-box. The second part of the annotation is optional, and consists in the manual correction of the predicted mask.
Multiple strategies are investigated to prioritize the correction of instances within a given training set. They are compared in terms of the annotation time budget required to achieve a same trained model accuracy. 
As a main contribution, our study reveals that only defining the extreme points and using the masks approximated by DEXTR to train the model reduces by up to a factor of 10 the annotation time required to train a model with same accuracy, but based on a fully-manual polygon-based delineation of the training instances. 
In addition, two original ordering metrics are proposed to prioritize the manual corrections of the approximated masks, so as to maximize the trained model accuracy profits. The prioritization of the corrections appears to lead to significant (up to $80\%$) savings in manual correction time, especially in the early stage of the correction process.

The rest of the paper is organized as follows. After a short survey of the related work in Section~\ref{sec:SoA}, our proposed method is introduced in Section~\ref{sec:methodology} and validated in Section~\ref{sec:results}.

\section{Related Work}
\label{sec:SoA}


\subsection{Instance Annotation}

Instance masks are traditionally generated manually, by drawing polygons around instances \cite{dutta2019vgg}.
Tools like Graph Cut\cite{boykov2001interactive} and GrabCut\cite{rother2004grabcut} have rapidly been considered to turn a bounding-box into a reasonably accurate segmentation mask, to reduce the manual intervention to bounding box definition and mask correction. Further assistance is now provided to annotators in the form of deep learning (DL) tools that typically refine a prior information, consisting in a coarse and incomplete annotation provided in the form of a bounding box \cite{deformablegrid, CurveGCN2019, acuna2018efficient, scribblebox}, an approximate contour \cite{Coarse2Fine,Acuna_2019_CVPR}, or a set of points lying on the instance border \cite{Man+18,CurveGCN2019,acuna2018efficient,Wang_2019_CVPR, benenson2019large}.

Those methods all aim at defining the mask of an instance at minimal human annotation load. For all of them, the human annotation cost is split into a prior annotation cost, and a mask correction cost. 
Similarly, our work adopts the Deep Extreme Cut (DEXTR) method \cite{Man+18} to turn a manually defined prior into an approximated instance mask. DEXTR has been chosen because it relies on a cheap prior, consisting in extreme points of an instance (left-most, right-most, top, bottom pixels), and because it is representative of modern DL-based annotation assistance. Despite it also leverages some prior information to predict an approximate mask, our work differs from previous art because, given a set of images, it is primarily interested in studying how to best allocate a human annotation time budget to the prior definition and to the correction of specific instance masks, so as to maximize the accuracy of the instance segmentation model trained from the resulting annotated data. This question is more general than simply estimating the time savings when generating a same set of accurate training masks since it also considers the benefit brought by approximated masks to the trained model accuracy.

Deciding to which instances the human annotation resources should be allocated first can be envisioned as an active learning problem, in which a learning algorithm interactively queries the human annotator to label well-chosen samples with the desired outputs.
In the context of semantic segmentation, several previous work have already proposed to select additional images to annotate according to a metric derived from the model trained based on already annotated samples. 
For example, Yang et al. \cite{yang2017suggestive} propose to select the biological images to annotate in decreasing order of trained network prediction uncertainty, as estimated based on bootstrapping \cite{EfroTibs93}.
A similar strategy has been used successfully in the medical field, e.g. Gorriz et al. \cite{Gorriz2017CostEffectiveAL}, Kuo et al. \cite{kuo2018cost} or Zhang et al. \cite{Zhang_2020}, and is shown to achieve good semantic segmentation performance with a smaller number of annotated samples.
In an instance segmentation context, the work in \cite{IoUpred_2020_MTA} proposes to select the image samples to annotate in priority by estimating the intersection-over-union between the predicted and true instance mask. This estimation requires the tedious training of an additional neural network branch.
In contrast, our work leverages the prior information derived from instance extreme points to decide which instance masks should be corrected in priority.

As a valuable complement to our work, which aims at providing a training set that is sufficient to train a reasonably accurate model at low manual cost, a very recent work~\cite{scalingup_2021_ICCV} has focused on bridging the gap between a preliminary model and the large training set required to train a highly accurate model~\cite{dataneed1,dataneed2}. Given a model trained based on a relatively small set of manually annotated images and a large set of unlabeled images, the goal of~\cite{scalingup_2021_ICCV} is to populate the unlabeled set with high-quality segmentation masks using as little human intervention as possible. This work comes thus downstream to our method, and give additional value to our study since it demonstrates that high-quality masks can be obtained almost for free as long as images and a reasonably accurate initial model are available.

\subsection{Mixed annotations qualities}

The impact of using datasets with heterogeneous label quality when training a segmentation CNN has been studied in recent years \cite{ImportanceLabelQuality, ke2020learning}. The main conclusion was that using accurate labels for only a small fraction of the data, the rest being weakly annotated, does not lead to a huge drop of performances in segmentation quality ( Ke et al. \cite{ke2020learning}). Zlateski et al. \cite{ImportanceLabelQuality} also concluded that when annotation time is an issue, gathering many coarse annotations is more important than producing a few fine ones. They advised to spend as much time for both coarse and fine levels, which implies that having a larger number of simpler labels than the amount of fine labels still leads to satisfactory performances. Our work extends those investigations to the instance segmentation case. The main difference between semantic and instance segmentation lies in the fact that (i) instance mask can be reasonably be approximated based on the automatic refinement of a prior information (like extreme points), and (ii) the annotation of instance borders might require a high precision to train the model correctly, especially when this border lies between two overlapping instances. Hence, how the annotation load/precision should be spread among instances is worth investigating. This question is central to our paper and to the lessons drawn from our experiments. 

\subsection{Adapting training to weak supervision.}
Instead of relying on instance masks to supervise the training of a segmentation model, some recent works have introduced losses that directly supervise the training based on the instance bounding boxes. Therefore, Tian et al. \cite{tian2020conditional,tian2020boxinst} propose to combine a projection loss, which compares the predicted mask to the ground truth bounding box, with an affinity loss that promotes similar instance labels for neighboring pixels. These solutions however still fall short compared to strong supervision. By mixing DEXTR and manually corrected masks, our solution allows to trade-off the relative importance of weak and strong supervision.

\begin{figure*}
\centering
  \includegraphics[width=.78\linewidth]{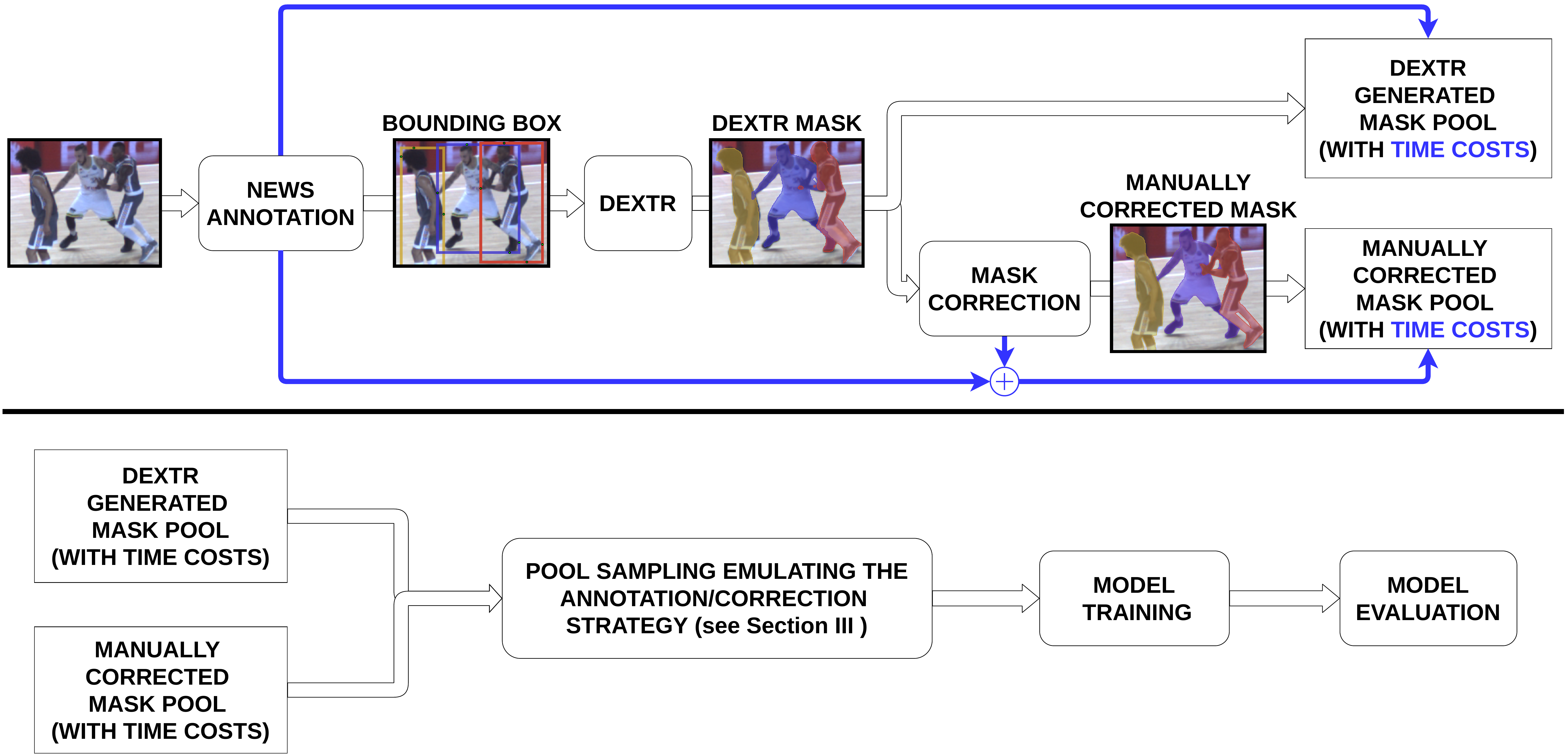}
  \caption{ 
    Our framework to study the trade-off between annotation load and trained model quality: [Top] Annotation flowchart: North/East/West/South extreme points are defined for every object instance in input image, and associated time cost is recorded (blue arrows). Masks are then generated from those keypoints using an off-the-shelf segmentation model, i.e. DEXTR \cite{Man+18}, to fill a pool of so-called DEXTR masks. Those masks are manually corrected, and associated time cost recorded, to fill a pool of ground-truth instance mask annotations. [Bottom] The two pools of instance masks are sampled to mimic an annotation strategy of interest, up to the depletion of a manual annotation time budget. An instance segmentation model is trained based on the resulting annotations. Its performance on a test set, for which ground-truth masks are available, measures the trained model quality corresponding to the annotation strategy and time budget of interest.
    }
    \label{fig:annotation_pipeline}
\end{figure*}

\section{Methodology}
\label{sec:methodology}

The purpose of our study is to derive a number of guidelines regarding the creation of a dataset dedicated to the training of an instance segmentation model for a specific application. Images of instances-of-interest are assumed to be available, and the recommended guidelines target a cost-effective use of human resources when defining the instance segmentation masks required to train the model. Hence, those guidelines aim at maximizing the quality of the automatic instance segmentation carried out by the learnt model, for a given amount of human-resources involved in the annotation of training images. Our work considers two approaches to locate instances in a training image. The first one consists in defining the extreme points of an object instance (left-most, right-most, top, bottom pixels). Those extreme points define the object bounding box, and are referred to as NEWS (North/East/West/South) keypoints in the following. In contrast, the second approach approximates the shape of the object with one polygon or more to handle occlusions. The number of polygons per instance, and edges/vertices in each polygon are not fixed a priori. They depend on the shape complexity and can be manually controlled through the annotation interface (see supplementary material for GUI user guide).
\noindent

The investigation conducted in this paper is depicted in Figure~\ref{fig:annotation_pipeline}, and builds (i) on the observation that the annotation of polygon shapes is significantly more time-consuming than the manual definition of the NEWS keypoints of an instance (see Section \ref{subsec:exp-config}), and (ii) on the recent success of works that have proposed to learn generic CNN models to transform the rough information provided by NEWS keypoints into a dense segmentation mask of an object that matches those extreme points \cite{Man+18,Wang_2019_CVPR}. Those two facts led to the annotation flowchart depicted in Figure~\ref{fig:annotation_pipeline} (top), which leverages bounding box knowledge and DEXTR to approximate the instance mask.

Our work compares different ways of assigning manual annotation resources to instances, to delineate their shape with a manually defined polygon, or to define their NEWS keypoints and, optionally, to correct the mask predicted by DEXTR from the bounding box prior. For each strategy, the resulting masks, corrected or not, are used to supervise the training of the Panoptic-Deeplab instance segmentation CNN architecture \cite{cheng2020panopticdeeplab}, and the comparison between strategies assesses the trade-off between the learned model quality (quantified based on test-time segmentation quality metric) and the manual annotation load (measured in units of time allocated to the annotation). 
We now present the different annotation strategies compared in the rest of the paper.

\subsection{Frame by frame annotation}
\label{subsec:image-strategy}

This section considers that the training images are annotated one after the other in a random order, up to the depletion of the manual annotation time budget. 
The prefix FbF is used to refer to this 'frame by frame' annotation schedule. When adopting this schedule, the same annotation strategy is followed for all images. Three different strategies are envisioned. First, the manual annotation of polygons is considered as a baseline, and is denoted {\bf FbF-M}, where M stands for 'Manual'.
The second strategy proposes to define the NEWS keypoints for every instance in the image, and relies on DEXTR \cite{Man+18} to turn the extreme points into an instance mask. No manual correction of the resulting mask is considered by this strategy, and it is denoted {\bf FbF-BB}, where FbF stands for 'Frame-by-Frame', while BB denotes 'Bounding Box'.
The third strategy, denoted {\bf FbF-BB+C}, completes the second one by manually correcting the instance mask approximated by DEXTR. 

\noindent
The human resources required by each annotation strategy increases with the number $N$ of annotated frames, as well as the learned model quality. Hence, for each strategy, by changing $N$, the learned model quality can be plotted as a function of the involved manual resources, i.e. of the manual annotation time. These plots are presented in Section~\ref{sec:results}.

\subsection{Frame by frame correction}
\label{subsec:image-correction}

Our experiments have revealed that the manual annotation cost associated with the definition of extreme NEWS keypoints is relatively small compared to the cost of correcting the mask predicted by DEXTR. Therefore, we propose to first annotate the NEWS keypoints on the entire training dataset, before correcting the masks predicted from those NEWS keypoints on a frame-by-frame basis. In practice, we consider two variants of the frame by frame correction strategy. Those variants are denoted {\bf BB4All-FC} where 'BB4All' indicates that bounding boxes have been defined for all dataset instances before starting the correction of DEXTR masks, and the letters in 'FC', refer to 'frame by frame' (F), and 'correction' (C). 'FC' is thus used to specify that all instances are corrected in the frame subject to correction. Note that, even if BB4All-FC processes the dataset images frame by frame, the BB4All schemes fundamentally differ from the FbF strategies defined above. This is because BB4All assumes that the instance bounding boxes have been manually defined for all images in the dataset, before starting the DEXTR mask correction. Hence, even in absence of correction, BB4All can use the entire set of images, and their approximated instance annotations, as training set. In contrast, FbF starts from an empty training set that is progressively augmented with new images.

\begin{figure}[ht]
\centering
\includegraphics[width = .25\textwidth]{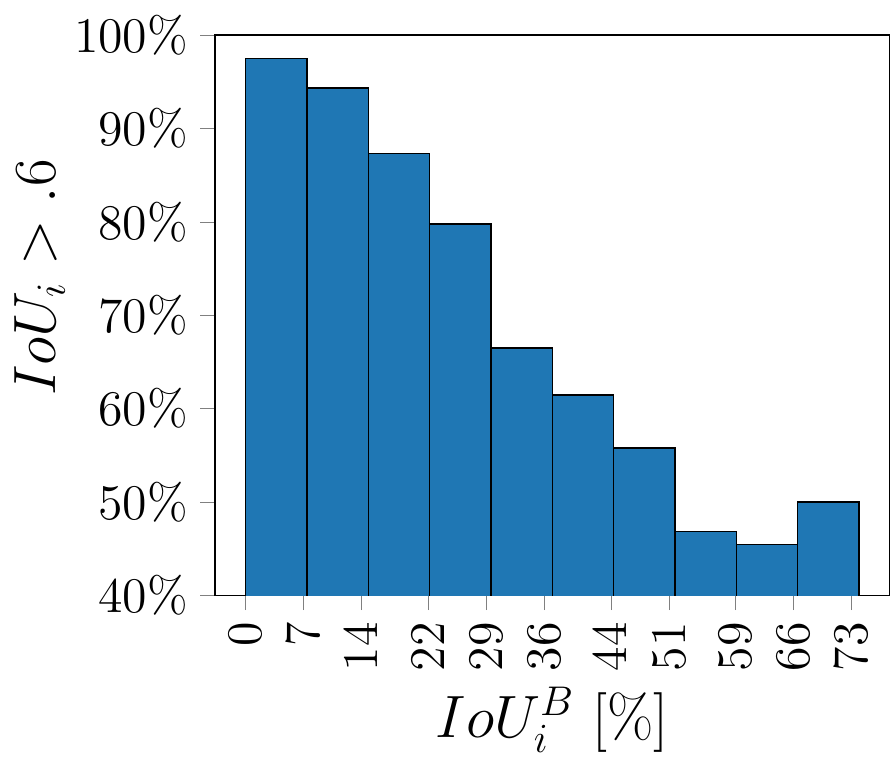}
   \caption{ Percentage of instances that are correctly segmented from their NEWS keypoints, i.e. having $IoU_{i} > 0.6$, as a function of their maximal bounding box overlap with another instance, as measured in bins of $IoU^B_i$ values. The more an instance is isolated from others (i.e. low $IoU^{B}_i$ bins), the more likely its DEXTR mask is correct, i.e. $IoU_i > 0.6$. This confirms the relevance of prioritizing the correction of overlapping instances.  }
   \label{fig:IoU_M_dextrVScorr}
\end{figure}

\subsection{Ordering the instance correction}
\label{subsec:instance-strategy}

As an alternative to the frame by frame correction of instance masks presented in Section~\ref{subsec:image-correction}, this section proposes to select the masks to correct in priority independently of the image they belong to. The NEWS keypoints are still assumed to have been defined for all instances in the dataset, and two different strategies are considered to define the order in which DEXTR mask should be corrected. 

\medskip 
\subsubsection{Prioritizing instances with overlapping bounding boxes}

Ideally, instances that are likely to be poorly segmented by the automatic DEXTR-based transformation of their extreme points should be corrected first. In practice, the segmentation quality of an instance can be measured by the intersection-over-union ($IoU$) between its estimated and ground-truth mask. Formally, for the $i^{th}$ instance, we have
{\small
\begin{equation}
IoU_{i} = \frac{M_{i}^{GT} \cap M_{i}}{M_{i}^{GT} \cup M_{i}},
\end{equation}
}
where $M_i$ and $M_{i}^{GT}$ denote the DEXTR and corrected mask of the $i^{th}$ instance, respectively.
An instance is generally considered to be correctly segmented when its $IoU_{i} > 0.6$.
As depicted in Figure~\ref{fig:IoU_M_dextrVScorr}, the proportion of instances that are correctly segmented increases in case of small overlap between instance bounding-boxes. 
In other words, the chance that an instance is poorly segmented increases with the maximal overlap between the instance bounding box and the bounding box of another instance in the image. As a consequence, correcting the instance masks in decreasing order of this overlap is expected to correct in priority the instance masks that are likely to be subject to the largest errors. Formally, let $\cal{I}$ denote the set of instances in a given dataset, and $B_i$ denote the bounding box of instance $i \in \cal{I}$. We introduce $\text{IoU}^\text{B}_i$ to stand for the maximal intersection-over-union between the bounding boxes of instance $i$ and any other instance $k$ in $\cal{I}$. Mathematically,
{\small
\begin{equation}
   \text{IoU}^\text{B}_i = \max_{k\in \cal{I}} \frac{B_i \cap B_k}{B_i \cup B_k}.
\end{equation}
}
A larger bounding-box overlap $\text{IoU}^\text{B}_i$ increases the chance of occlusions between instance $i$ and another instance, and makes errors in the automatic transformation of the extreme points by DEXTR more likely. The strategy that corrects instance masks in decreasing order of bounding-box overlap is referred to as {\bf BB4All-IC-Oo}, where 'IC' stands for 'Instance Correction, while 'Oo' corresponds to 'Overlap-based ordering'.

\medskip 
\subsubsection{Prioritizing based on the active learning paradigm}

Extreme points are again assumed to have been annotated for all instances, but we consider that an instance segmentation model has been trained based on the approximated masks predicted from those NEWS keypoints. 

Following the active learning principle, this model provides a prior information that can be used to select the mask instances to correct in priority as the ones for which there is a large mismatch between the mask predicted by DEXTR and the one predicted by the model trained from the whole set of uncorrected masks\footnote{Note that our AL strategy only relies on the model trained from the whole set of uncorrected masks, as directly predicted by DEXTR from the instance bounding boxes, and does not update this model as more masks have been corrected.  Preliminary experiments have indeed revealed that such update does not improve the ordering of subsequent corrections in terms of the benefit they bring to subsequently trained models.}. Formally, let $M_i$ denote the mask predicted by DEXTR from the extreme points of instance $i$, and ${\cal{M}}_i$ denotes the set of masks predicted by the trained model on the image containing instance $i$. We introduce $\text{IoU}^*_i$ to denote the maximal intersection-over-union between the mask $M_i$ and any other mask $M$ in ${\cal{M}}_i$. Mathematically,
{\small
\begin{equation}
   \text{IoU}^*_i = \max_{M \in {\cal{M}}_i} \frac{M_i \cap M}{M_i \cup M}.
\end{equation}
}
Since a small value of $\text{IoU}^*_i$ for instance $i$ reflects a large discrepancy between the mask predicted by the trained model and the mask $M_i$ obtained from the extreme points, it becomes relevant to correct (or validate) $M_i$ in increasing order of $\text{IoU}^*_i$ values. 
When $\text{IoU}^*_i$ gets close to one, the masks derived from the extreme points and the trained model are consistent, and $\text{IoU}^*_i$ does not help in identifying instances to correct in priority. In that case, prioritizing the correction of overlapping instances, as explained above, remains valid. 

Hence, a meaningful prioritization should account both for the bounding box overlap, and for the mismatch between the DEXTR mask and the mask predicted by the trained model. Formally, let $c_i$ define the prediction confidence level of the $i^{th}$ instance as
{\small
\begin{equation}
   c_i = \text{IoU}^*_i + \alpha . \text{IoU}^*_i . ( 1- \text{IoU}^\text{B}_i),
  \label{eq:confidence}
\end{equation}
}

with $\alpha$ defining the importance of $\text{IoU}^\text{B}_i$ in the relative priority of instances. In practice, $\alpha$ has been set to 0 or 1 in our experiments. The strategy that selects the instances to correct in increasing order of confidence levels is referred to as the {\bf BB4All-IC-ALo} strategy, with 'IC' standing for 'Instance Correction', and 'ALo' refering to 'Active Learning ordering'. Variants of the strategy described above could obviously be defined, e.g. by training a more accurate instance segmentation model once a fraction of instances have been corrected. However, in practice, the largest benefit obtained from this strategy has been observed when prioritizing the first corrections, generally corresponding to the larger and most impactful errors among DEXTR-based masks.

\section{Experiments}
\label{sec:results}

This section aims at comparing the annotation strategies presented in Section~\ref{sec:methodology} in terms of the trade-off between trained model quality and manual annotation load. 
It first introduces the experimental set-up, as well as the training quality metrics. It then plots the trained model quality as a function of the manual annotation load, for various annotation strategies. Recommendations regarding ways to annotate a novel image dataset in a cost-effective manner are formulated based on the analysis of those plots. 

\subsection{Experimental set-up and assessment metrics}
\label{subsec:exp-config}

{\bf Use case.} Our experiments aim at studying the creation of a dataset associated with a reasonably challenging and specific instance segmentation use case. Therefore, a set of basket-ball game images has been considered. Using a dedicated interface implemented with nodeJS/Typescript\footnote{See\cite{DeepsportDataset} for the GUI user guide, describing the interface functionalities.}, NEWS extreme points have been manually defined for all player instances in the dataset, and their corresponding DEXTR masks have been automatically generated and manually corrected. Polygons have also been manually defined to delineate the player instances, to provide a reference baseline to which methods based on NEWS keypoints can be compared, both in terms of time and accuracy. The time associated with those two manual steps has been recorded, 
using the \texttt{ts-stopwatch} library. A group of six annotators has been involved in the annotation and correction process.
On average, NEWS keypoints are defined in $4$ s/instance, while the correction of the masks requires $45$ or $70$ s/instance, depending on whether the instance NEWS-based bounding box is isolated or partly overlaps another instance bounding box. The average time for a fully manual annotation of polygons is 95 s/instance. The dataset, its corresponding annotations, and the user guide associated with the interface are available at \cite{DeepsportDataset}.
This dataset consists of 561 images from 26 arenas involving a large variety of lighting conditions. Each image captures one half court with a resolution between 2Mpx and 5Mpx. 100 images have been extracted for the testing such that games in the test set come from arenas that are not considered during training. In final, around 3900 instances were used for training, distributed among 461 images and around 800 instances were used for testing, distributed among the 100 test images.

{\bf Instance segmentation CNN.} Each annotation strategy and annotation time budget results in a specific training set, made of a fraction of the dataset instances and their segmentation mask (the manual one, the DEXTR one, or its corrected version). 
To compare the value of those training sets in terms of trained model quality, the Panoptic Deeplab \cite{cheng2020panopticdeeplab} has been selected as a reference instance segmentation CNN model, since it corresponds to a conceptually simple state-of-the-art solution to address instance segmentation tasks. In practice, given our limited computational resources\footnote{3 RTX 2080 TI 11GB, Intel(R) Xeon(R) CPU E5-1650 v4  12 cores @ 3.60GHz, 64 GB RAM}, this model has been used with a MobileNet backbone~\cite{sandler2018mobilenetv2}, a batch size equal to 12, and for 8000 iterations. We used the following losses: \textit{hard pixel mining} \cite{gu2020hard} as a semantic loss, \textit{mean squared error} as center loss, and finally the \textit{l1} metric as our offset loss. We used an \textit{Adam} solver and a linear learning rate scheduler combined with a base learning rate of $0.001$. All the models in our experiments were trained from weights that were pre-trained on or ImageNet\cite{ImageNet} and available on Pytorch website \cite{init_weights}.


{\bf Quality metric.} The trained models are assessed using the Panoptic Quality (PQ), as introduced in \cite{kirillov2019panoptic}. It corresponds to the product of the segmentation quality (SQ) and the recognition quality (RQ), defined by comparing the predicted masks with the ground-truth ones.


    
Specifically, a predicted instance mask $M^{*}$ is
identified as a True Positive mask (TP), if its IoU with one of the ground truth instance masks $M^{GT}_j$ is higher than the threshold of 0.5: 
        {\small
        \begin{equation}\label{eq:sq-tp}
        ( M^{*},\  M^{GT}_j) \in {{\rm TP}} \Leftrightarrow \exists j,~{\rm IoU}( M^{*},\  M^{GT}_j) \ge 0.5.
        \end{equation}
        }

Otherwise, the instance mask is considered as a False Positive (FP).    
The SQ is then defined as the averaged IoU over the TP pairs, namely
        {\small
        \begin{equation}\label{eq:metric-psq}
            {\rm SQ} \coloneqq {\frac{1}{|{\rm TP}|}} \sum_{(u,v) \in {\rm TP}}{\rm IoU}(u,v),
        \end{equation}
        }
 and the RQ is defined as the $F_1$-score, i.e.
        {\small
        \begin{equation}\label{eq:metric-pdq}
            {\rm RQ} \coloneqq \frac{2|{\rm TP}|}{|{\cal{M}}|+|M^{GT}|},
        \end{equation}
        }
with $\cal{M}$ denoting the set of all predicted instances, and $M^{GT}$ being the set of all instances in the ground truth.

\begin{figure}[ht!]
    \centering

    \includegraphics[width=.4\textwidth]{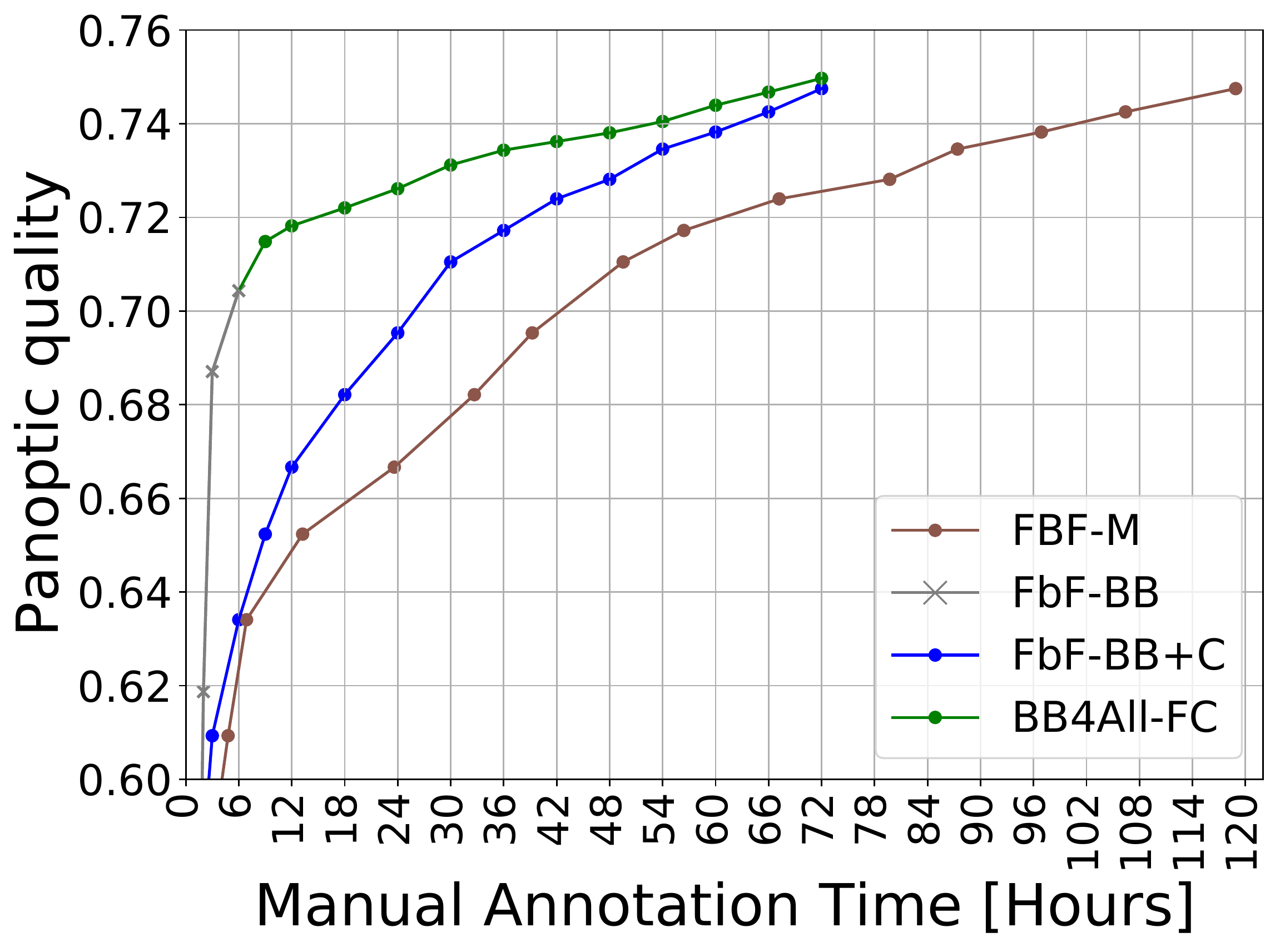}

    \caption{Panoptic Quality as a function of annotation time.
    The approximated masks generated by DEXTR from the NEWS keypoints (grey/green curve) achieve a reasonable panoptic quality with 8 to 10 times less human annotation resources than a fully manual annotation of the masks. Using DEXTR masks alone also reduces human annotation by a factor 5 compared to DEXTR followed by a manual correction of the approximated masks. See the text for the definition of methods.  }
    \label{fig:PQoverTime}
\end{figure}

\begin{figure}
    \centering

    \includegraphics[width=.4\textwidth]{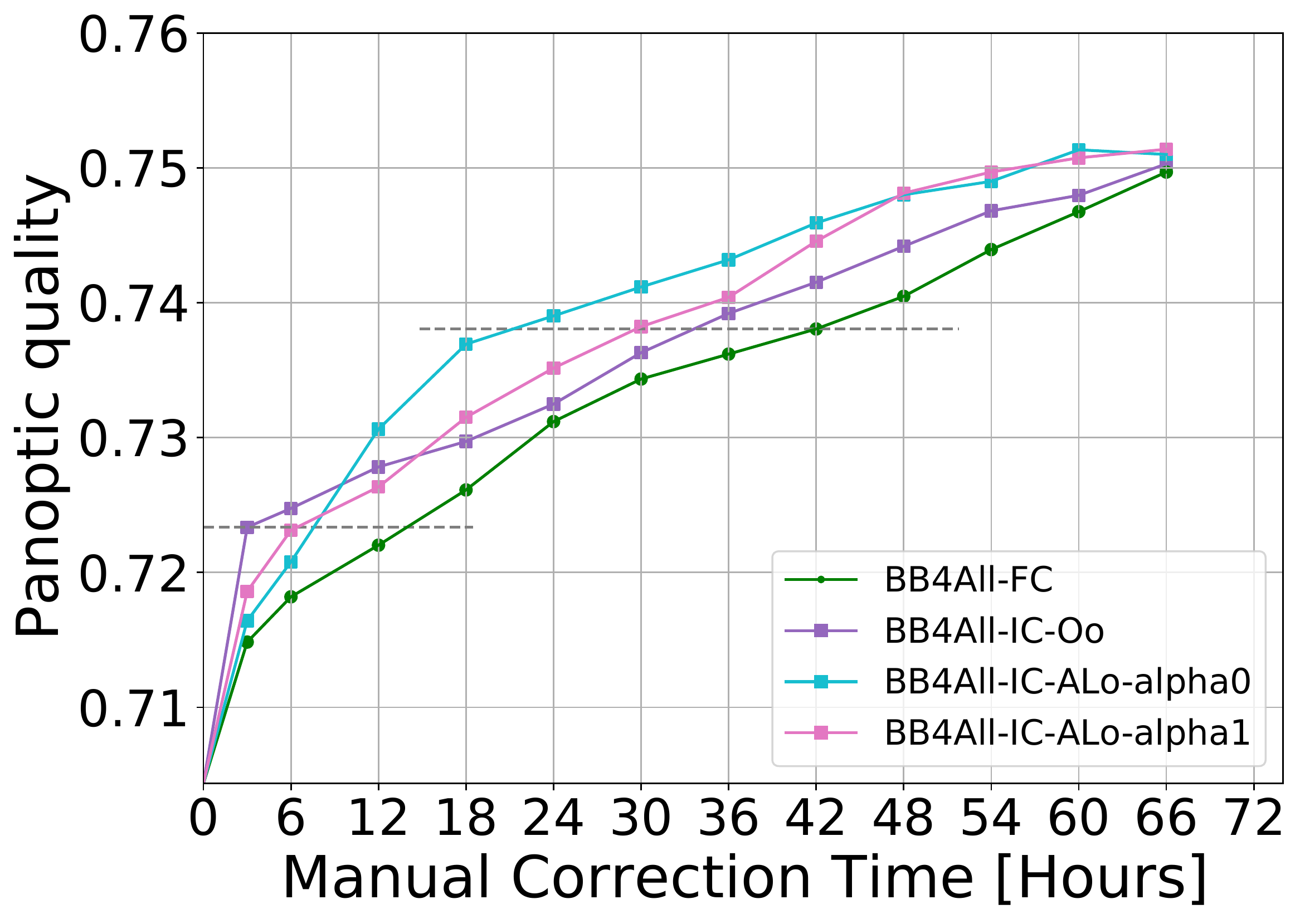}

    \caption{Panoptic Quality as a function of correction time, assuming the NEWS keypoints have been annotated for the whole dataset. Carefully ordering the correction of DEXTR masks is beneficial, especially in case of limited annotation resources. Prioritization is done either based on the bounding box overlap (Oo), or based on the mismatch between the masks predicted by DEXTR and by the model trained from the whole set of uncorrected masks (ALo). See the text for details. }
    \label{fig:PQoverCorrectionTime}
\end{figure}

\subsection{Trained model quality vs. annotation time}
\label{subsec:trainedquality}

Figure \ref{fig:PQoverTime} depicts, as a function of the annotation time, the panoptic quality of the Panoptic-Deeplab instance segmentation models trained with the datasets resulting from the various annotation strategies introduced in Section~\ref{sec:methodology}.  

In this figure, the FbF prefix indicates that the annotation is done frame by frame, in a random order. FbF-M defines the masks manually, and provides a reference baseline. The FbF-BB annotation strategy assigns the DEXTR mask (as predicted from the instance NEWS keypoints) to each instance in the frame to annotate, while FbF-BB+C considers the manually corrected version of this mask. By comparing those curves, we observe that, at $70\%$ PQ, the FbF-BB requires 5 hours of annotation, which is more than 8 times less human annotation resources than a fully manual delineation of instances with polygons. It is also 5 times less than the 25 hours required when correcting the DEXTR masks frame by frame, as done by FbF-BB+C. Interestingly, we also observe that, with the whole dataset, FbF-BB reaches up to near $71\%$ panoptic quality, which is less than $5\%$ below the quality achieved with fully corrected annotation. This is in line with other works \cite{ImportanceLabelQuality} that state that training with many samples whose annotation is prone to noise should be preferred to training with only few but perfectly-annotated ones. 
The first recommendation of our paper is thus to promote the coarse definition of instance masks, using DEXTR and NEWS keypoints, over the entire dataset before considering their progressive correction.
The rest of this section investigates how to prioritize this correction process.

\subsection{Prioritizing the correction of DEXTR masks.}

The first approach considered to define the order in which the masks should be corrected is based on the observation that instances that overlap each other are more delicate to segment from their extreme points (see Figure~\ref{fig:IoU_M_dextrVScorr}). 
This led to the strategy BB4All-IC-Oo, which computes the DEXTR masks for the whole training set, and then corrects those masks in decreasing order of overlap, as defined by $\text{IoU}^\text{B}_i$.



A second approach to define how to prioritize the correction of DEXTR masks is inspired by the active learning paradigm. 
Since we have observed in Figure~\ref{fig:PQoverTime} that FbF-BB achieves reasonably good performance at low annotation cost, we propose to use the Panoptic-Deeplab model trained based on the uncorrected DEXTR masks to predict instances in our training set, and select the DEXTR masks to correct in increasing order of prediction confidence, as defined in Equation (\ref{eq:confidence}). This strategy is denoted BB4All-IC-ALo, with 'ALo' referring to the active learning ordering principle.
Despite the Panoptic-Deeplab model is applied to the same images than the ones used for training, early stopping prevents overfitting and preserves the capacity to differentiate confident and unreliable masks. Early stopping was implemented by reducing the number of training epochs. 

Figure~\ref{fig:PQoverCorrectionTime} shows that carefully ordering the correction helps compared to a frame by frame correction. Prioritizing based on bounding box overlap provides the largest benefit when the manual correction time budget is limited. Slightly above $72\%$ PQ accuracy, it reduces the correction load by close to $80\%$ compared to the baseline frame by frame correction (3 hours of correction for BB4All-IC-Oo vs. 14 hours for BB4All-FC). At higher correction time budget, ordering the corrections based on the mismatch between DEXTR and a preliminary model (BB4All-IC-ALo-alpha0) performs best, saving $50\%$ of human correction resources (21 vs. 42 hours of correction) a bit below $74\%$ PQ accuracy.

\section{Conclusion}

Our work builds on a universal prior-based segmentation model to accelerate the annotation of instance masks in a dataset of images.
Experiments, run with DEXTR as a universal model and using extreme points as a prior, have shown that our solution leads to significant gains (up to 10 times smaller) in annotation time compared to a fully manual annotation. Our study has also revealed the benefit of generating the prior on the entire dataset before allocating the remaining annotation resources to the correction of masks predicted by DEXTR based on this prior. Eventually, the prioritization of the corrections appears to lead to significant (generally as high as $50\%$, with a peak reaching $80\%$) savings in manual correction time, especially in the early stage of the correction process. Our experiments thus demonstrate (i) the advantage of collecting bounding box priors for all instances before considering the correction of some of them, and (ii) the gain obtained when prioritizing the corrections.

\bibliographystyle{IEEEtran}
\bibliography{egbib}
%



\end{document}